\documentclass[11pt,a4paper]{article}
\usepackage{authblk}
\usepackage[hyperref]{naaclhlt2019}
\usepackage{times}
\usepackage{latexsym}
\usepackage{tabularx}
\usepackage{graphicx}
\usepackage{amssymb}
\usepackage{amsmath}
\usepackage{relsize}
\usepackage{paralist}
\usepackage{multirow}
\usepackage{stmaryrd}

\usepackage{url}

\aclfinalcopy 

\title{Text Generation from Knowledge Graphs with Graph Transformers}

\author[1]{Rik Koncel-Kedziorski}
\author[1]{Dhanush Bekal}
\author[1]{Yi Luan}
\author[2]{Mirella Lapata}
\author[1,3]{Hannaneh Hajishirzi}
\affil[1]{University of Washington}
\affil[ ]{\texttt{\{kedzior,dhanush,luanyi,hannaneh\}@uw.edu}}
\affil[2]{University of Edinburgh}
\affil[ ]{\texttt{mlap@inf.ed.ac.uk}}
\affil[3]{Allen Institute for Artificial Intelligence}

\date{}

\begin{document}
\maketitle
\begin{abstract}

Generating texts which express complex ideas spanning multiple sentences requires a structured representation of their content (document plan), but these representations are prohibitively expensive to manually produce. 
In this work, we address the problem of generating coherent multi-sentence texts from the output of an information extraction system, and in particular a knowledge graph. 
Graphical knowledge representations are ubiquitous in computing, but pose a significant challenge for text generation techniques due to their non-hierarchical nature, collapsing of long-distance dependencies, and structural variety. 
We introduce a novel graph transforming encoder which can leverage the relational structure of such knowledge graphs without imposing linearization or hierarchical constraints.
Incorporated into an encoder-decoder setup, we provide an end-to-end trainable system for graph-to-text generation that we apply to the domain of scientific text. 
Automatic and human evaluations show that our technique produces more informative texts which exhibit better document structure than competitive encoder-decoder methods. 
\footnote{Data and code available at \url{https://github.com/rikdz/GraphWriter}}
\end{abstract}

\section{Introduction}

\begin{figure}
    \centering
    \includegraphics[width=\columnwidth]{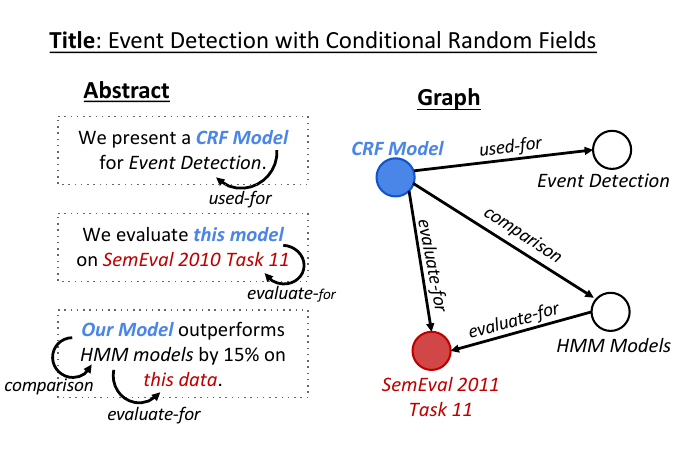}
    \caption{A scientific text showing the annotations of an information extraction system and the corresponding graphical representation. Coreference annotations shown in color. Our model learns to generate texts from automatically extracted knowledge using a graph encoder decoder setup.}
    \label{fig:teaser}
\end{figure}
Increases in computing power and model capacity have made it possible to generate mostly-grammatical sentence-length strings of natural language text.
However, generating several sentences related to a topic and which display overall coherence and discourse-relatedness is an open challenge.
The difficulties are compounded in domains of interest such as scientific writing. 
Here the variety of possible topics is great (e.g. topics as diverse as driving, writing poetry, and picking stocks are all referenced in one subfield of one scientific discipline). 
Additionally, there are strong constraints on document structure, as scientific communication requires carefully ordered explanations of processes and phenomena. 

Many researchers have sought to address these issues by working with structured inputs. 
Data-to-text generation models \cite{konstas-lapata:2013:EMNLP,lebret2016neural,wiseman2017challenges,puduppully2019data} condition text generation on table-structured inputs. 
Tabular input representations provide more guidance for producing longer texts, but are only available for limited domains as they are assembled at great expense by manual annotation processes. 

The current work explores the possibility of using information extraction (IE) systems to automatically provide context for generating longer texts (Figure~\ref{fig:teaser}). 
Robust IE systems are available and have support over a large variety of textual domains, and often provide rich annotations of relationships that extend beyond the scope of a single sentence.
But due to their automatic nature, they also introduce challenges for generation such as erroneous annotations, structural variety, and significant abstraction of surface textual features (such as grammatical relations or predicate-argument structure).

To effect our study, we use a collection of abstracts from a corpus of scientific articles \cite{ammar:18}.
We extract entity, coreference, and relation annotations for each abstract with a state-of-the-art information extraction system \cite{luan2018multi}, and represent the annotations as a {\it knowledge graph} which collapses co-referential entities. 
An example of a text and graph are shown in Figure~\ref{fig:teaser}.
We use these graph/text pairs to train a novel attention-based encoder-decoder model for knowledge-graph-to-text generation. 
Our model, GraphWriter, extends the successful Transformer for text encoding~\cite{vaswani2017attention} to graph-structured inputs, building on the recent Graph Attention Network architecture~\cite{velickovic2017graph}. 
The result is a powerful, general model for graph encoding which can incorporate global structural information when contextualizing vertices in their local neighborhoods. 

The main contributions of this work include:
\begin{compactenum}
    \item We propose a new graph transformer encoder that applies the successful sequence transformer to graph structured inputs. 
    \item We show how IE output can be formed as a connected unlabeled graph for use in attention-based encoders. 
    \item We provide a large dataset of knowledge-graphs paired with scientific texts for further study.
\end{compactenum}
Through detailed automatic and human evaluations, we demonstrate that automatically extracted knowledge can be used for multi-sentence text generation. 
We further show that structuring and encoding this knowledge as a graph leads to improved generation performance compared to other encoder-decoder setups. 
Finally, we show that GraphWriter's transformer-style encoder is more effective than Graph Attention Networks on the knowledge-graph-to-text task. 

\section{Related Work}



Our work falls under the larger scope of concept-to-text generation. 
\citet{barzilay2005collective} introduced a collective content selection model for generating summaries of football games from tables of game statistics. 
\citet{LiangJK:ACL09} jointly learn to segment and align text with records, reducing the supervision needed for learning. 
\citet{kim2010generative} improve this technique by learning a semantic parse to logical forms. 
\citet{konstas-lapata:2013:EMNLP} focus on the generation objective, jointly learning planning and generating using a rhetorical (RST) grammar induction approach. 

These earlier works often focused on smaller record generation datasets such as WeatherGov and RoboCup, but recently \citet{mei2016talk} showed how neural models can achieve strong results on these standards, prompting researchers to investigate more challenging domains such as ours.

\citet{lebret2016neural} tackles the task of generating the first sentence of a Wikipedia entry from the associated infobox. 
They provide a large dataset of such entries and a language model conditioned on tables.
Our work focuses on a multi-sentence task where relations can extend beyond sentence boundaries. 

\citet{wiseman2017challenges} study the difficulty of applying neural models to the data-to-text task. 
They introduce a large dataset where a text summary of a basketball game is paired with two tables of relevant statistics and show that neural models struggle to compete with template based methods over this data. 
We propose generating from graphs rather than tables, and show that graphs can be effectively encoded to capture both local and global structure in the input. 

We show that modeling knowledge as a graph improves generation results, connecting our work to other graph-to-text tasks such as generating from Abstract Meaning Representation (AMR) graphs.
\citet{Konstas2017NeuralAS} provide the first neural model for this task, and show that pretraining on a large dataset of noisy automatic parses can improve results.
However, they do not directly model the graph structure, relying on linearization and sequence encoding instead.
Current works improve this through more sophisticated graph encoding techniques. 
\citet{marcheggiani2018deep} encode input graphs directly using a graph convolution encoder \cite{Kipf2016SemiSupervisedCW}.
Our model extends the graph attention networks of \citet{velickovic2017graph}, a direct descendant of the convolutional approach which offers more modeling power and has been shown to improve performance.
\citet{Song2018AGM} uses a graph LSTM model to effect information propagation. 
At each timestep, a vertex is represented by a gated combination of the vertices to which it is connected and the labeled edges connecting them. 
\citet{beck2018graph} use a similar gated graph neural network. 
Both of these gated models make heavy use of label information, which is much sparser in our knowledge graphs than in AMR. 
Generally, AMR graphs are denser, rooted, and connected, whereas the knowledge our model works with lacks these characteristics. 
For this reason, we focus on attention-based models such as \citet{velickovic2017graph}, which impose fewer constraints on their input. 

Finally, our work is related to \citet{wang2018paper} who offer a method for generating scientific abstracts from titles. 
Their model uses a gated rewriter network to write and revise several draft outputs in several sequence-to-sequence steps.
While we operate in the same general domain as this work, our task setup is ultimately different due to the use of extracted information as input.
We argue that our setup improves the task defined in \citet{wang2018paper}, and our more general model can be applied across tasks and domains. 

\section{The AGENDA Dataset}
\begin{table}[t]
\begin{center}
    \begin{tabular}{lccc}
         & Title & Abstract & KG  \\ \hline
        Vocab & 29K  & 77K  & 54K  \\
        Tokens & 413K  & 5.8M & 1.2M \\
        Entities & - & - & 518K \\ 
        Avg Length & 9.9  & 141.2  & -  \\
        Avg \#Vertices & - & - & 12.42 \\
        Avg \#Edges &  - & - & 4.43 
        \end{tabular}
    \caption{Data statistics of our AGENDA dataset. Averages are computed per instance. }
    \label{tab:stats}
    \end{center}
\end{table}

We consider the problem of generating a text from automatically extracted information ({\it knowledge}). 
IE systems can produce high quality knowledge for a variety of domains, synthesizing information from across sentence and even document boundaries. 
Generating coherent text from knowledge requires a model which considers global characteristics of the knowledge as well as local characteristics of each entity. 
This feature of the task motivates our use of graphs for representing knowledge, where neighborhoods localize important information and paths through the graph build connections between distant nodes through intermediate ones.
An example knowledge graph can be seen in Figure~\ref{fig:teaser}.

We formulate our problem as follows: given the title of a scientific article and a knowledge graph constructed by an automatic information extraction system, the goal is to generate an abstract that a) is appropriate for the given title and b) expresses the content of the knowledge graph in natural language text. 
To evaluate how well a model accomplishes this goal, we introduce the Abstract GENeration DAtaset (AGENDA), a dataset of knowledge graphs paired with scientific abstracts.
Our dataset consists of 40k paper titles and abstracts from the Semantic Scholar Corpus taken from the proceedings of 12 top AI conferences \cite{ammar:18}. 

For each abstract, we create a knowledge graph in two steps. 
First, we apply the SciIE system of \citet{luan2018multi}, a state-of-the-art science-domain information extraction system. 
This system provides named entity recognition for scientific terms, with entity types Task, Method, Metric, Material, or Other Scientific Term. 
The model also produces co-reference annotations as well as seven relations that can obtain between different entities (Compare, Used-for, Feature-of, Hyponym-of, Evaluate-for, and Conjunction).
For example, in Figure~\ref{fig:teaser}, the node labeled ``SemEval 2011 Task 11'' is of type `Task', ``HMM Models'' is of type `Model', and there is a `Evaluate-For' relation showing that the models are evaluated on the task. 

We form these annotations into knowledge graphs. 
We collapse co-referential entities into a single node associated with the longest mention (on the assumption that these will be the most informative). 
We then connect nodes to one another using the relation annotations, treating these as labeled edges in the graph. 
The result is a possibly unconnected graph representation of the SciIE annotations for a given abstract.

Statistics of the AGENDA dataset are available in Table~\ref{tab:stats}.
We split the AGENDA dataset into 38,720 training, 1000 validation, and 1000 test datapoints.
We offer standardized data splits to facilitate comparison. 

\section{Model}

\begin{figure}
    \centering
    \includegraphics[width=\columnwidth]{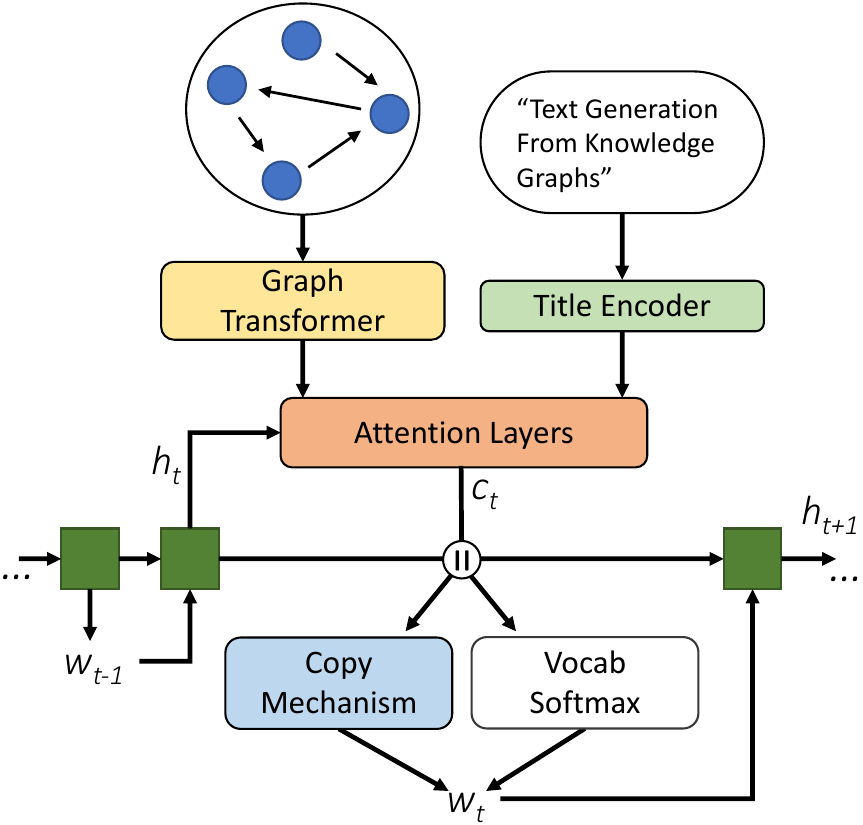}
    \caption{GraphWriter Model Overview}
    \label{fig:model}
\end{figure}

Following most work on neural generation we adopt an encoder-decoder architecture, shown in Figure~\ref{fig:model}, which we call GraphWriter.
The input to GraphWriter is a title and a knowledge graph which are encoded respectively with a bidirectional recurrent neural network and a novel Graph Transformer architecture (to be discussed in
  Section~\ref{sec:encoder}). 
At each decoder time step, we attend on encodings of the knowledge graph and document title using the decoder hidden state $\mathbf{h}_t \in \mathbb{R}^{d}$.
The resulting vectors are used to select output $w_t$ either from the decoder's vocabulary or by copying an entity from the knowledge graph. 
Details of our decoding process are described in Section~\ref{sec:decode}.
The model is trained end-to-end to minimize the negative log likelihood of the mixed copy and vocabulary probability distribution and the human authored text.

\subsection{Encoder}\label{sec:encoder}

\begin{figure}[t]
    \centering
    \includegraphics[width=.8\columnwidth]{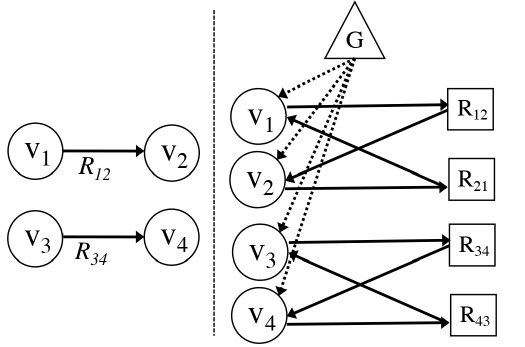}
    \caption{Converting disconnected labeled graph to connected unlabeled graph for use in attention-based encoder. $v_i$ refer to vertices, $R_{ij}$ to relations, and $G$ is a global context node.}
    \label{fig:graph}
\end{figure}

The AGENDA dataset contains a knowledge graph for each datapoint, but our model requires unlabeled, connected graphs as input. 
To encode knowledge graphs with this model, we restructure each graph as an unlabeled connected graph, preserving label information by the method described below and sketched in Figure~\ref{fig:graph}. 

\paragraph{Graph Preparation}
We convert each graph to an unlabeled connected bipartite graphs following a similar procedure to \citet{beck2018graph}. 
In this process, each labeled edge is replaced with two vertices: one representing the forward direction of the relation and one representing the reverse.
These new vertices are then connected to the entity vertices so that the directionality of the former edge is maintained.
This restructures the original knowledge graph as an unlabeled directed graph where all vertices correspond to entities and relations in the SciIE annotations without loss of information. 
To promote information flow between disconnected parts of the graph, we add a global vertex which connects all entity vertices. 
This global vertex will be used to initialize the decoder, analogously to the final encoder hidden state in a traditional sequence to sequence model.
The final result of these restructuring operations is a connected, unlabeled graph $G =(V,E)$, where $V$ is a list of entities, relations, and a global node and $E$ is an adjacency matrix describing the directed edges.

\begin{figure}
    \centering
    \includegraphics[width=\columnwidth]{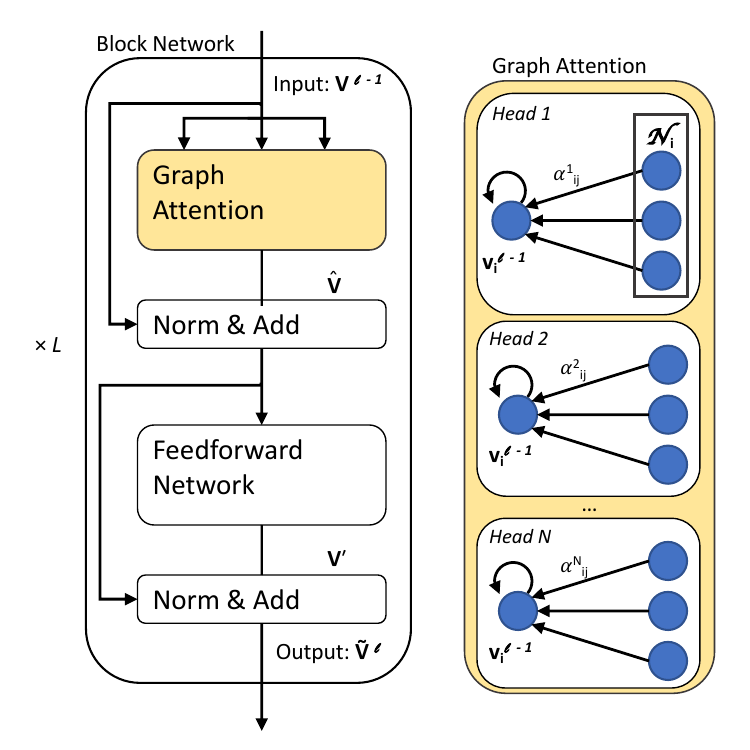}
    \caption{Graph Transformer}
    \label{fig:enc}
\end{figure}
\paragraph{Graph Transformer} 
Our model is most similar to the Graph Attention Network (GAT) of \citet{velickovic2017graph}, which computes the hidden representations of each node in a graph by attending over its neighbors following a self-attention strategy.
The use of self-attention in GAT addresses the shortcomings of prior methods based on graph convolutions \cite{Defferrard2016ConvolutionalNN,Kipf2016SemiSupervisedCW}, but limits vertex updates to information from adjacent nodes. 
Our model allows for a more global contextualization of each vertex through the use of a transformer-style architecture. 
The recently proposed Transformer \cite{vaswani2017attention} addresses the inherent sequential computation shortcoming of recurrent neural networks, enabling efficient and paralleled computation by invoking a self-attention mechanism for global context modeling.
These models have shown promising results in a variety of text processing tasks \cite{radford2018improving}.

Our Graph Transformer encoder starts with self-attention of local neighborhoods of vertices; the key difference with GAT is that our model includes additional mechanisms for capturing global context. 
This additional modeling power allows the Graph Transformer to better articulate how a vertex should be updated given the content of its neighbors, as well as to learn global patterns of graph structure relevant to the model's objective.

Specifically, $V$ is embedded in a dense continuous space by the embedding process described at the end of this section, resulting in matrix $\mathbf{V}^0=[\mathbf{v}_i], \mathbf{v}_i \in \mathbb{R}^{d}$ which will serve as input to the graph transformer model shown in Figure~\ref{fig:enc}.
Each vertex representation $\mathbf{v}_i$ is contextualized by attending over the other vertices to which $v_i$ is connected in $G$.  
We use an $N$-headed self attention setup, where $N$ independent attentions are calculated and concatenated before a residual connection is applied:
\begin{eqnarray}\label{eq:attn}
    \mathbf{\hat{v}}_i &=& {\bf v}_i + \bigparallel_{n=1}^N \sum_{j \in \mathcal{N}_i}\alpha^n_{ij} \mathbf{W}^n_{V} \mathbf{v}_j \\
     \alpha^n_{ij} &=& a^n({\bf v}_i,{\bf v}_j)
\end{eqnarray}
Here, $\|$ denotes the concatenation of the $N$ attention heads, $\mathcal{N}_i$ denotes the neighborhood of $v_i$ in $G$, $\mathbf{W}^n_{V} \in \mathbb{R}^{(\frac{d}{N})\times d}$, and where $a^n$ are attention mechanisms parameterized per head.
In this work, we use attention functions of the  following form:
\begin{equation}
    a({\bf q}_i,{\bf k}_j) = \frac{\exp(({\mathbf{W}_{K}{\bf k}_j)^{\top}\mathbf{W}_{Q}\mathbf{q}_i})}{\sum_{z \in \mathcal{N}_i}\exp(({\mathbf{W}_{K}{\bf k}_z)^{\top}\mathbf{W}_{Q}\mathbf{q}_i})}
\end{equation}
Each $a$ learns independent transformations $\mathbf{W}_{Q}, \mathbf{W}_{K} \in \mathbb{R}^{(\frac{d}{N})\times d}$ of ${\bf q}$ and ${\bf k}$ respectively, and the resulting product is normalized across all connected edges. 
To reduce the tendency of these dot products to impede gradient flow, we scale them by $\frac{1}{\sqrt{d}}$, following \citet{vaswani2017attention}.

The Graph Transformer then augments these  multi-headed attention layers with {\it block} networks. 
Each block applies the following transformations: 
\begin{eqnarray}
    \mathbf{\tilde{v}}_i &=& \textrm{LayerNorm}(\mathbf{v}'_i + \textrm{LayerNorm}(\mathbf{\hat{v}}_i))\\
    \mathbf{v}'_i &=& \textrm{FFN}(\textrm{LayerNorm}(\mathbf{\hat{v}}_i))
\end{eqnarray}
Where FFN$(\mathbf{x})$ is a two layer feedforward network with a non-linear transformation $f$ between layers i.e. $f( \mathbf{x}\mathbf{W}_1+b_1)\mathbf{W}_2+b_2$. 

Stacking multiple blocks allows information to propagate through the graph.
Blocks are stacked $L$ times, with the output of layer $l-1$ taken as the input to layer $l$, so that $\mathbf{v}^{l}_i = \mathbf{\tilde{v}}^{l-1}_i$.
The resulting vertex encodings $\mathbf{V}^L = [{\bf v}^L_i]$ represent entities, relations, and the global node contextualized by their relationships in the graph structure. 
We refer to the resulting encodings as {\it graph contextualized vertex encodings}.

\paragraph{Embedding Vertices, Encoding Title}
As stated above, the vertices of our graph correspond to entities and relations from the SciIE annotations.
Because each relation is represented as both a forward- and backward-looking vertex, we learn two embeddings per relation as well as an initial embedding for the global node.
Entities correspond to scientific terms which are often multi-word expressions. 
To produce a single $d$-dimensional embedding per phrase, we use the last hidden state of a bidirectional RNN run over embeddings of each word in the entity phrase, i.e. $\textrm{BiRNN}({\bf x}_1 \ldots {\bf x}_m)$ for dense embeddings ${\bf x}$ and phrase length $m$. 
The output of our embedding step is a collection ${\bf V}^0$ of $d$-dimensional vectors representing each vertex in $V$.

The title input is also a short string, and so we encode it with another BiRNN to produce ${\bf T} = \textrm{BiRNN}(x'_1 \ldots x'_m)$ for title word embedding ${\bf x'}$.

\subsection{Decoder}\label{sec:decode}
We decode with an attention-based decoder with a copy mechanism for copying input from the knowledge graph and title. 
At each decoding timestep $t$ we use decoder hidden state ${\bf h}_t$ to compute context vectors ${\bf c}_g$ and ${\bf c}_s$ for the graph and title sequence respectively.
${\bf c}_g$ is computed using multi-headed attention contextualized by ${\bf h}_t$:
\begin{eqnarray}\label{eq:attn2}
    \mathbf{c}_g &=& \mathbf{h}_t + \bigparallel_{n=1}^N \sum_{j \in V}\alpha^n_{j} \mathbf{W}^n_{G} \mathbf{v^L}_j \\
    \alpha_{j} &=& a(\mathbf{h}_t,\mathbf{v^L}_j)
\end{eqnarray}
for $a$ as described in Equation~(\ref{eq:attn}) by attending over the graph contextualized encodings  ${\bf V}^L$.
${\bf c}_s$ is computed similarly, attending over the title encoding ${\bf T}$.  
We then construct the final context vector by concatenation, ${\bf c}_t = [{\bf c}_g\|{\bf c}_s]$.
We use an input-feeding decoder \cite{Luong2015EffectiveAT} where both ${\bf h}_t$ and ${\bf c}_t$ are passed as input to the next RNN timestep. 

 We compute a probability $p$ of copying from the input using ${\bf h}_t$ and ${\bf c}_t$ in a fashion similar to \citet{see2017get}, that is:
 \begin{equation}
     p = \sigma ( \mathbf{W}_{copy} [{\bf h}_t\|{\bf c}_t] + b_{copy} )
 \end{equation}
 The final next-token probability distribution is:
\begin{equation}
p * \alpha^{copy} + (1-p) * \alpha^{vocab}, 
\end{equation}
Where the probability distribution $\alpha^{copy}$ over entities and input tokens is computed as $\alpha^{copy}_j = a([{\bf h}_t\|{\bf c}_t],\mathbf{x}_j)$ for ${\bf x}_j \in {\bf V}\|{\bf T}$.
 The remaining $1-p$ probability is given to $\alpha^{vocab}$, which is calculated by scaling $[{\bf h}_t\|{\bf c}_t]$ to the vocabulary size and taking a softmax.

\section{Experiments}

\paragraph{Evaluation Metrics} \label{sec:eval}
We evaluate using a combination of human and automatic evaluations. 
For \textbf{human} evaluation, participants were asked to compare abstracts generated by various models and those written by the authors of the scientific articles. 
We used Best-Worst Scaling (BWS; \cite{louviere1991best,louviere2015best}), a less labor-intensive alternative to paired comparisons that has been shown to produce more reliable results than rating scales \cite{Kiritchenko2016CapturingRF}.
Participants were presented with two or three abstracts and asked to decide which one was better and which one was worse in order of grammar and fluency (is the abstract written in well-formed English?), coherence (does the abstract have an introduction, state the problem or task, describe a solution, and discuss evaluations or results?), and informativeness (does the abstract relate to the provided title and make use of appropriate scientific terms?).
We provided examples of good and bad abstracts and explain how they succeed or fail to meet the defined criteria. 

Because our dataset is scientific in nature, evaluations must be done by experts and we can only collect a limited number of these high quality datapoints.\footnote{Attempts to crowd source this evaluation failed.}
The study was conducted by 15 experts (i.e. computer science students) who were familiar with the abstract writing task and the content of the abstracts they judged. 
To supplement this, we also provide \textbf{automatic} metrics. 
We use BLEU \cite{Papineni2002BleuAM}, an n-gram overlap measure popular in text generation tasks, and METEOR \cite{Denkowski2014MeteorUL}, a machine translation with paraphrase and language-specific considerations.

\paragraph{Comparisons} 
We compare our GraphWriter against several strong baselines. 
In GAT, we replace our Graph Transformer encoder with a Graph Attention Network of \citep{velickovic2017graph}.
This encoder consists of PReLU activations stacked between 6 self-attention layers.
To determine the usefulness of including graph relations, we compare to a model which uses only entities and title (EntityWriter).
Finally, we compare with the gated rewriter model of \citet{wang2018paper} (Rewriter).
This model uses only the document title to iteratively rewrite drafts of its output. \footnote{Due to the larger size and greater variety of our dataset and accompanying vocabularies compared to theirs, we were unable to train this model with the reported batch size of 240. We use batch size 24 instead, which is partially responsible for the lower performance.}

\paragraph{Implementation Details}
Our models are trained end-to-end to minimize the negative joint log likelihood of the target text vocabulary and the copied entity indices. 
We use SGD optimization with momentum \cite{Qian1999OnTM} and ``warm restarts'', a cyclical regiment that reduces the learning rate from 0.25 to 0.05 over the course of 5 epochs, then resets for the following epoch.
Models are trained for 15 epochs with early stopping \cite{Prechelt:1998} based on the validation loss, with most models stopping between 8 and 13 epochs. 
We use single-layer LSTMs \cite{Hochreiter:1997} as recurrent networks.
We use dropout \cite{Srivastava:2014} in self attention layers set to 0.3.
Hidden states and embedding dimensions are fixed at 500 and attentions learn 500 dimensional projections.
In Block layers, the feedforward network has an intermediate size of 2000, and we use a PReLU activation function \cite{he2015delving}.
GraphWriter and GAT use $L=6$ layers. 
The number of attention heads is set to 4. 
In all models, for both inputs and output, we replace words occurring fewer than 5 times with $<${\it unk}$>$ tokens. 
In each abstract, we replace all mentions in a coreference chain in the abstract with the canonical mention used in the graph.
We decode with beam search \cite{graves2012sequence,sutskever2014sequence} with a beam size of 4. 
A post-processing step deletes repeated sentences and repeated coordinated clauses.

\subsection{Results} 
A comparison of all systems in terms of automatic metrics is shown in Table~\ref{tab:compare}.
Our GraphWriter model outperforms other methods.
We see that models which leverage title, entities, and relations (GraphWriter and GAT) outperform models which use less information (EntityWriter and Rewriter). 

We see that GraphWriter outperforms GAT across metrics, indicating that the global contextualization provided by GraphWriter improves generation. 
To verify the performance gap between GraphWriter and GAT, we report the average test metrics for 4 training runs of each model along with their variances. 
We see that the variance of the different models is non-overlapping, and in fact all training runs of GraphWriter outperformed all runs of GAT on these metrics. 

\begin{table}\begin{small}
\begin{center}
\begin{tabular}{lcc}
     & BLEU & METEOR \\ \hline
    GraphWriter  & {\bf 14.3} $\pm$ 1.01 & {\bf 18.8} $\pm$ 0.28  \\
    GAT & 12.2 $\pm$ 0.44 & 17.2 $\pm$ 0.63  \\
    EntityWriter & 10.38 & 16.53 \\ 
    Rewriter & 1.05 & 8.38 \\ \hline
\end{tabular}
\caption{Automatic Evaluations of Generation Systems.}
\label{tab:compare}
\end{center}\end{small}
\end{table}

\paragraph{Does Knowledge Help?}
To evaluate the value of knowledge in the generation task we compare our GraphWriter model to a model which does not generate from knowledge.  
We provide expert annotators with 50 randomly-selected paper titles from the test set and ask them for a single judgment according to the criteria described in Section~\ref{sec:eval}. 
We pair each paper title with the generated abstracts produced by GraphWriter (a knowledge-informed modes), Rewriter (a knowledge-agnostic model), and the gold abstract (with canonicalized coreferential mentions).

Results of this comparison can be seen in Table~\ref{tab:know}. 
We see that GraphWriter is selected as ``Best'' more often than Rewriter, and is less often selected as ``Worst'', attesting to the value of including knowledge in the text generation process.
We see that sometimes generated texts are preferred to human authored text, which is due in part to the disfluencies introduced by canonicalization of entity mentions. 

\begin{table}
\begin{small}
\begin{center}
\begin{tabular}{lcc}
     & Best & Worst\\ \hline
    Rewriter (No knowledge) & 12\% & 64\%\\
    GraphWriter (Knowledge) & 24\% & 36\% \\
    Human Authored & 64\% & 0\% \\ \hline
\end{tabular}
\caption{Does knowledge improve generation? Human evaluations of best and worst abstract.}
\label{tab:know}
\end{center}
\end{small}
\end{table}
\begin{table}
\begin{small}
\begin{center}
\begin{tabular}{lrrr}
     & Win & Lose & Tie \\ \hline
    Structure  & 63\%  & 17\% & 20\%  \\
    Informativeness & 43\% & 23\% & 33\%   \\
    Grammar & 63\% & 23\% & 13\% \\ 
    Overall &  63\%  & 17\% & 20\% \\ \hline
\end{tabular}
\caption{Human Judgments of GraphWriter and EntityWriter models.}
\label{tab:VsEnt}
\end{center}
\end{small}
\end{table}

\begin{table*}[t]
\begin{small}
\begin{tabularx}{\textwidth}{lX}
    Title & Block and Group Regularized Sparse Modeling for Dictionary Learning   \\ 
    Knowledge & (dictionary learning, CONJUNCTION, sparse coding) ; (optimization problems, USED-FOR, dictionary learning) ; (optimization problems, USED-FOR, sparse coding)\ldots \\ \hline \\
    GraphWriter & Sparse representations have recently been shown to be effective in many optimization problems. However, existing dictionary learning methods are limited in the number of dictionary blocks, which can be expensive to obtain. In this paper, we propose a novel approach to dictionary learning based on sparse coding  \ldots\\
    GAT &  In this paper, we consider the problem of dictionary learning in well-known datasets. In particular, we consider the problem of dictionary learning, where the goal is to find a set of dictionary blocks that maximize the likelihood of a given set of dictionary blocks \ldots \\
    EntityWriter & We propose a novel dictionary learning framework for reconstructed block/group sparse coding schemes. The dictionary learning framework is based on the descent, which is a block structure of the group structure  \ldots\\ 
    Rewriter & This paper presents a new approach to the k-means of the algorithm. The proposed approach is based on the basis of the stationarity algorithm. The algorithm is based on the fact that the number of bits is a constant of the base of the base of the input  \ldots\\
    {\it Gold} & This paper proposes a dictionary learning framework that combines the proposed block/group (BGSC) or reconstructed block/group (R-BGSC) sparse coding schemes with the novel Intra-block Coherence Suppression Dictionary Learning algorithm. An important and distinguishing feature of the proposed framework is that all dictionary blocks are trained simultaneously \ldots \\ \hline \hline \\
    Title & Image Interpolation with Directionlets \\
    Knowledge & (directionally adaptive image interpolation USED-FOR edge information) ; (numeric and visual quality, HYPONYM-OF, directionally adaptive image interpolation) ; (directionlets, EVALUATE-FOR, multiple-direction wavelet transform) \ldots \\ 
    \hline \\
GraphWriter & In this paper, we propose a novel directionally adaptive image interpolation based on the multiple-direction wavelet transform, called directionlets, which can be used as a directionlets to improve the numeric and visual quality of the directionally adaptive image interpolation \ldots \\
GAT & In this paper, we propose a novel directionally adaptive image interpolation, called directionally adaptive image interpolation, for directionally adaptive image interpolation , which is based on the multiple-direction wavelet transform \ldots \\
EntityWriter & We present a novel directionally adaptive image interpolation for numeric and visual quality. The wavelet transform is based on the wavelet transform between the low-resolution image and the interpolated image. The high-resolution image is represented by a wavelet transform \ldots \\
Rewriter & We present a new method for finding topic-specific data sets. The key technical contributions of our approach is to be a function of the terrestrial distributed memory. The key idea is to be a function of the page that seeks to be ranked the buckets of the data. The basic idea is a new tool for the embedded space \ldots \\
{\it Gold} & We present a novel directionally adaptive image interpolation based on a multiple-direction wavelet transform, called directionlets. The directionally adaptive image interpolation uses directionlets to efficiently capture directional features and to extract edge information along different directions from the low-resolution image \ldots
\end{tabularx}
\end{small}
\caption{Example outputs of various systems versus Gold.}
\label{tab:output}
\end{table*}
To further understand the advantages of using knowledge graphs, we provide a more detailed comparison of the GraphWriter and EntityWriter models.
We select 30 additional test datapoints and ask experts to provide per-criterion judgments of the outputs of the two systems. 
Since both models make use of extracted entities, we show this list along with the title for each datapoint, and modify the description of Informativeness to include ``making use of the provided entities''.
Results of this evaluation are shown in Table~\ref{tab:VsEnt}.
Here we see that including structured knowledge in the form of a graph improves abstract generation compared to generating from an unstructured collection of entities.  
The largest gains are made in terms of document structure and grammar, indicating that the structure of the input knowledge is being translated into the surface form.




\paragraph{Generating from Title} 
The Rewriter model \cite{wang2018paper} considers the task of generating an abstract with only the paper's title as input. 
We compare against this model because it is among the first end-to-end systems to attempt to write scientific abstracts. 
However, the task setup used in \citet{wang2018paper} differs significantly from the task introduced in this work.
In order to make a fair comparison, we construct a variant of our model which is only provided with a title as input. We develop a model that predicts entities from the title, and then uses our knowledge-aware model to generate the abstract. 
For this comparison we use the EntityWriter model with a collection of entities inferred from the title alone (InferEntityWriter).

To infer relevant entities, we learn to embed titles and entities extracted from the corresponding abstract in a shared dense vector space by minimizing their cosine distance.
We use negative sampling to provide definition to this vector space. 
At test time, we use the title embedding to infer the $K=12$ closest entities to feed into the InferEntityWriter model. 
Results are shown in Table~\ref{rewritvsseq}, which shows that InferEntityWriter achieves  better results than Rewriter, indicating that the intermediate entity prediction step is helpful in abstract generation. 

\begin{table} \begin{small}
\begin{center}
\begin{tabular}{lcc}
     & BLEU & METEOR \\ \hline
    Rewriter & 1.05 & 8.38 \\
    InferEntityWriter & 3.60 & 12.2  \\ \hline
\end{tabular}
\caption{Comparison of generation without knowledge and with Inferred Knowledge (InferEntityWriter)}
\label{rewritvsseq}
\end{center}\end{small}
\end{table}

\subsection{Analysis}

Table~\ref{tab:output} shows examples of various system outputs for a particular test instance.
We see that GraphWriter makes use of more entities from the input, arranged with more articulated textual context. 
It demonstrates less repetition than GAT. 
Both GraphWriter and GAT show much better coherence than EntityWriter, which copies entities from the input into unreasonable contexts. 
Rewriter, while fluent and grammatical, jumps from topic to topic, failing to relate as strongly to the input as the knowledge-aware models.


To determine the shortcomings of our model, we calculate rough error statistics over the outputs of the GraphWriter on the test set. 
We notice that 40\% of entities in the knowledge graphs do not appear in the generated text. 
Future work should address this coverage problem, perhaps through modifications to the inference procedure or a coverage loss \citep{tu2016modeling} modified to the specifics of this task. 
%
We find that 18\% of all sentences generated by our model repeat sentences or clauses and are subjected to the post-processing pruning mentioned in Section~\ref{sec:eval}.
While this step is a simple solution to improve generated outputs, a more advanced solution is required. 

\section{Conclusion}    
We have studied the problem of generating multi-sentence text from the output of automatic information extraction systems, and have shown that incorporating knowledge as graphs improves performance.
We introduced GraphWriter, featuring a new attention model for graph encoding, and demonstrated its utility through human and automatic evaluation compared to strong baselines.
Lastly, we provide a new resource for the generation community, the AGENDA dataset of abstracts and knowledge. 
Future work could address the problem of repetition and entity coverage in the generated texts. \
\section*{Acknowledgments}
This research was supported by the Office of Naval Research under the MURI grant N00014-18-1-2670, NSF (IIS 1616112, III 1703166), Allen Distinguished Investigator Award, Samsung GRO and gifts from Allen Institute for AI, Google, Amazon, and Bloomberg. We gratefully acknowledge the support of the European Research Council (Lapata; award number 681760). We also thank the anonymous reviewers and the UW-NLP group for their helpful comments.

\bibliography{refs}
\bibliographystyle{acl_natbib}

\end{document}


\section{Inferring Entities}\label{appendix}

For corroborating the fact that including any form of knowledge improves the quality of generated outputs, we compare model [ref] with the rewriter[]. In this comparison setup, we do not provide gold entities to our model. We instead provide list of retrieved entities which are closely related to the title at hand.

Entity retrieval systems can be formulated in multiple ways. Literature shows tasks such as graph completion, entity retrieval and entity recommendation. We develop a simple setup for the task of retrieving closely related entities to a given scientific title. We train a cosine embedding loss based system which minimizes the distance between a title and related entities and maximizes the distance between the title and negative samples of entities.

When a test title is fed to this system, we extract entities in the test title, perform jaccard distance comparison with entities in training titles and retrieve training titles with with Jaccard distance > 0.7. This Jaccard distance measure hints that the two titles contain similar information. We then extract entities present in the abstracts of those training titles and run them through the cosine similarity model together with the test title. The system then outputs a subset of entities which have the highest cosine similarity with the test title. Based on the entity statistics shown in table []. We retrieve utmost 12 closest entities. Note: If no entity is present in the test title, we extract subspans of length 1-4 for the jaccard distance comparison.

This list of inferred entities and the title are fed to our trained generation model [ref]. We compare the automatic metrics of this system with that of the rewriter [] where no entity information is available. We can clearly see that the system performs much better when some information is provided.

\section{Sample Outputs}

\begin{table*}[htb]
\renewcommand{\tabularxcolumn}{m} 
\small
\begin{tabularx}{\textwidth}{|c|>{\raggedright}X|}
\hline
Inputs & \textbf{Title :}  A Bayesian Model for Simultaneous Image Clustering , Annotation and Object Segmentation . \\ \textbf{Entities :} heterogeneous mix of components ; spatially contiguous objects ; logistic stick-breaking process ; non-parametric bayesian model ; variational bayesian analysis ; object-feature mixture models ; image classes ; mixture models ; image databases ; image features ; object types ; inference \\ \textbf{Relations :} image databases - used-for - inference ; logistic stick-breaking process - used-for - spatially contiguous objects ; variational bayesian analysis - used-for - inference \\ \tabularnewline
\hline \hline

Graph Writer & in this paper , we address the problem of inference from image databases . we propose a non-parametric bayesian model for learning a non-parametric bayesian model of a set of spatially contiguous objects , which are learned from data in a logistic stick-breaking process . we show that the logistic stick-breaking process can be learned from data , and can be learned from data in a logistic stick-breaking process . we demonstrate the effectiveness of our model on a set of image databases , and show that the logistic stick-breaking process is able to recover spatially contiguous objects from image databases . we also show that the logistic stick-breaking process can be used to perform inference and inference from image databases . \\ \tabularnewline
\hline

GAT Writer & in this paper , we propose a novel method for variational bayesian analysis , which can be used in a non-parametric bayesian model for inference . our approach is based on the logistic stick-breaking process , which can be used in a non-parametric bayesian model to learn heterogeneous mix of components from image databases . the proposed method can be applied to a variety of object types , such as spatially contiguous objects , in a logistic stick-breaking process . we show that the proposed method can be applied to other heterogeneous mix of components , such as variational bayesian analysis , which can be applied to image databases . we demonstrate the effectiveness of the proposed method by comparing its performance with the state-of-the-art methods on several image databases . \\ \tabularnewline
\hline

Entity Writer & we present a non-parametric bayesian model for object-feature mixture models that is based on a logistic stick-breaking process . the non-parametric bayesian model is a non-parametric bayesian model that is trained to predict the object types of a logistic stick-breaking process . we show that this non-parametric bayesian model can be used to predict the object types of a person from image databases . we show that the non-parametric bayesian model can be used to predict the object types of a person from image databases . \\ \tabularnewline
\hline

Rewriter & the problem of designing swarm is an information-theoretic approach to estimate the dominant electrode modulator from a given input . we propose a new approach for the problem of finding the largest unknown 2,1-norm of random projections . the algorithm is based on the solvability of the problem of finding a subset of the same algorithm . we propose a new approach for solving this problem , which is based on the solvability of the problem . the main idea is to devise a swarm optimization algorithm for solving the problem of finding the largest lyapunov matrices . the proposed algorithm is based on the same algorithm , and it is easy to implement the algorithm . we call this technique is based on a new type of algorithm , and show that the algorithm is able to achieve good performance in the presence of the noise . \\ \tabularnewline
\hline

Human & a non-parametric bayesian model is proposed for processing multiple images . the analysis employs image features and , when present , the words associated with accompanying annotations . the non-parametric bayesian model clusters the images into classes , and each image is segmented into a set of objects , also allowing the opportunity to assign a word to each object -lrb- localized labeling -rrb- . each object is assumed to be represented as a heterogeneous mix of components , with this realized via mixture models linking image features to object types . the number of image classes , number of object types , and the characteristics of the object-feature mixture models are inferred nonparametrically . to constitute spatially contiguous objects , a new logistic stick-breaking process is developed . inference is performed efficiently via variational bayesian analysis , with example results presented on two image databases . \\ \tabularnewline
\hline

\end{tabularx}
\end{table*}

\begin{table*}[htb]
\renewcommand{\tabularxcolumn}{m} 
\small
\begin{tabularx}{\textwidth}{|c|>{\raggedright}X|}
\hline
Inputs & \textbf{Title :}  
OFDM for underwater acoustic communications : Adaptive synchronization and sparse channel estimation .
\\ \textbf{Entities :} non-uniform frequency offset compensation ; phase synchronization method ; channel impulse response ; low-complexity channel estimation ; time domain ; decision-directed operation ; adaptive synchronization \\ \textbf{Relations :} adaptive synchronization - used-for - decision-directed operation
  \\ \tabularnewline
\hline \hline

Graph Writer & in this paper , we consider the problem of low-complexity channel estimation in the presence of channel impulse response . we propose a new phase synchronization method , which is based on the phase synchronization method , which is shown to be a phase synchronization method for the decision-directed operation . we show that the proposed phase synchronization method is robust to the decision-directed operation , and can be used to estimate the decision-directed operation . we also present a phase synchronization method for the decision-directed operation , which is shown to be robust to the channel impulse response . we also show that the proposed phase synchronization method can be used to improve the performance of the decision-directed operation . \\ \tabularnewline
\hline

GAT & in this paper , we consider the problem of low-complexity channel estimation in the presence of channel impulse response . in particular , we consider the problem of estimating the channel impulse response of a channel impulse response , where the channel impulse response is based on the estimation of the channel impulse response . we propose a new phase synchronization method which is based on the phase synchronization method . simulation results are provided to demonstrate the effectiveness of the proposed phase synchronization method . \\ \tabularnewline
\hline

Entity Writer & in this paper , we propose a phase synchronization method for low-complexity channel estimation . the phase synchronization method is based on the channel impulse response and is shown to be more robust than the phase synchronization method . the phase synchronization method is based on a phase synchronization method which is used to estimate the channel impulse response . the phase synchronization method is then used to estimate the channel impulse response of the channel impulse response . the phase synchronization method is applied to low-complexity channel estimation .  \\ \tabularnewline
\hline

Rewriter & we present a new method for the cake cutting traveling maximization of the input . the key technical contributions of this work is to develop a new type of vol-umetric functions that is based on the solvability of the '' codeword of '' economy '' . the key idea is to devise a new type of '' scattering '' that is guaranteed to converge the optimal weight of the optimal optimal solution . the main contribution is a function of the \ true '' '' product '' that is a simple and elegant algorithm for solving the problem of finding a divisible n-dimensional . the key to this basic approach is shown to be a new algorithm that is guaranteed to converge the number of dimensions . the key to this approach is shown to be a good approximation to the optimal algorithm . the key to this is that the algorithm is that the basic algorithm is easy to show that the algorithm is a simple and straightforward algorithm . \\ \tabularnewline
\hline

Human & a phase synchronization method , which provides non-uniform frequency offset compensation needed for wideband ofdm -lsb- 1 -rsb- , is coupled with low-complexity channel estimation in the time domain . sparsing of the channel impulse response leads to an improved performance , while adaptive synchronization supports decision-directed operation and yields low overhead . system performance is demonstrated using experimental data transmitted over a 1 km shallow water channel in the 19 khz-31 khz band . \\ \tabularnewline
\hline

\end{tabularx}
\end{table*}

\begin{table*}[htb]
\renewcommand{\tabularxcolumn}{m} 
\small
\begin{tabularx}{\textwidth}{|c|>{\raggedright}X|}
\hline
Inputs & \textbf{Title :}  Image interpolation with directionlets .
 \\ \textbf{Entities :} directionally adaptive image interpolation ; numeric and visual quality ; sharpness of details ; multiple-direction wavelet transform ; low-resolution image ; interpolated image ; high-resolution image ; edge information ; directional features ; directionlets \\ \textbf{Relations :} directionally adaptive image interpolation - used-for - edge information ; numeric and visual quality - hyponym-of - directionally adaptive image interpolation ; directionlets - evaluate-for - multiple-direction wavelet transform ; directionlets - used-for - directional features ; directionally adaptive image interpolation - used-for - directional features ; directionlets - used-for - directionally adaptive image interpolation ; multiple-direction wavelet transform - used-for - directionally adaptive image interpolation
 \\ \tabularnewline
\hline \hline

Graph Writer & in this paper , we propose a novel directionally adaptive image interpolation based on the multiple-direction wavelet transform , called directionlets , which can be used as a directionlets to improve the numeric and visual quality of the directionally adaptive image interpolation . in the proposed directionally adaptive image interpolation , directionally adaptive image interpolation is used to extract directional features from the interpolated image . the directionally adaptive image interpolation is based on the multiple-direction wavelet transform , which is a directionlets , and the sharpness of details is computed by using the multiple-direction wavelet transform of the interpolated image . the proposed directionally adaptive image interpolation is based on the multiple-direction wavelet transform , i.e. , the sharpness of details of the interpolated image , and the sharpness of details of the interpolated image is obtained . the proposed directionally adaptive image interpolation has been applied to both numeric and visual quality and the sharpness of details . \\ \tabularnewline
\hline

GAT Writer & in this paper , we propose a novel directionally adaptive image interpolation , called directionally adaptive image interpolation , for directionally adaptive image interpolation , which is based on the multiple-direction wavelet transform . the directionally adaptive image interpolation is based on the multiple-direction wavelet transform , which is an extension of the multiple-direction wavelet transform to the multiple-direction wavelet transform . the directionally adaptive image interpolation is based on the use of a multiple-direction wavelet transform , called the multiple-direction wavelet transform , which is based on the multiple-direction wavelet transform . the proposed multiple-direction wavelet transform can be applied to directionally adaptive image interpolation , which can be used to improve the performance of the multiple-direction wavelet transform . the experimental results show that the proposed directionally adaptive image interpolation significantly improves the performance of the multiple-direction wavelet transform , and that the proposed directionally adaptive image interpolation can achieve better performance than the state-of-the-art directionlets . \\ \tabularnewline
\hline

Entity Writer & we present a novel directionally adaptive image interpolation for numeric and visual quality . the wavelet transform is based on the wavelet transform between the low-resolution image and the interpolated image . the high-resolution image is represented by a wavelet transform , which is then used to estimate the sharpness of details . the high-resolution image is then used to estimate the sharpness of details . the high-resolution image is used to estimate the sharpness of details and the edge information . the high-resolution image is used to estimate the sharpness of details .  \\ \tabularnewline
\hline

Rewriter & we present a new method for finding topic-specific data sets . the key technical contributions of our approach is to be a function of the terrestrial distributed memory . the key idea is to be a function of the page that seeks to be ranked the buckets of the data . the basic idea is a new tool for the embedded space . the key idea is to be the main purpose of the problem of finding a few possible . the key idea is that the method is based on a series of parallel projections . the key idea is to be used to simplify the error-prone objective function to the problem of finding a quantum on a given input space . the key to this approach is based on a large scale of the '' interleaver '' of the beam of the data . the basic idea of the proposed method is demonstrated on a number of the puzzle of the algorithm . \\ \tabularnewline
\hline

Human & we present a novel directionally adaptive image interpolation based on a multiple-direction wavelet transform , called directionlets . the directionally adaptive image interpolation uses directionlets to efficiently capture directional features and to extract edge information along different directions from the low-resolution image . then , the high-resolution image is generated using this information to preserve sharpness of details . our directionally adaptive image interpolation outperforms the state-of-the-art methods in terms of both numeric and visual quality of the interpolated image . \\ \tabularnewline
\hline

\end{tabularx}
\end{table*}